%% file: main.tex
\documentclass[runningheads]{llncs}

\usepackage{eccv}

\usepackage{eccvabbrv}

\usepackage{graphicx}
\usepackage{booktabs}
\usepackage{makecell}
\usepackage{multirow}

\usepackage[accsupp]{axessibility}  

\usepackage{hyperref}

\usepackage{orcidlink}
\usepackage{wrapfig}

\begin{document}

\title{Unveiling and Mitigating Memorization in Text-to-image Diffusion Models through \\ Cross Attention} 

\titlerunning{Memorization and Cross Attention in Text-to-image Diffusion Models}

\author{Jie Ren\inst{1}\orcidlink{0000-0003-2663-6405} \and
Yaxin Li\inst{1}\orcidlink{0000-0002-6227-7844} \and
Shenglai Zeng\inst{1} \and
Han Xu\inst{1}\orcidlink{0000-0002-4016-6748} \and
Lingjuan Lyu\inst{2} \and \\
Yue Xing\inst{1}\orcidlink{0000-0001-7723-0048} \and
Jiliang Tang\inst{1}
}

\authorrunning{J.~Ren et al.}

\institute{Michigan State University \\
\email{\{renjie3,liyaxin1,zengshe1,xuhan1,xingyue1,tangjili\}@msu.edu} \and
Sony AI\\
\email{lingjuan.lv@sony.com}
}

\maketitle

\begin{abstract}
Recent advancements in text-to-image (T2I) diffusion models have demonstrated their remarkable capability to generate high-quality images from textual prompts. However, increasing research indicates that these models memorize and replicate images from their training data, raising concerns about potential copyright infringement and privacy risks. In our study, we provide a novel perspective to understand this memorization phenomenon by examining its relationship with cross-attention mechanisms. We reveal that during memorization, the cross-attention tends to focus disproportionately on the embeddings of specific tokens. The diffusion model is overfitted to these token embeddings, memorizing corresponding training images. To elucidate this phenomenon, we further identify and discuss various intrinsic findings of cross-attention that contribute to memorization. Building on these insights, we introduce an innovative approach to detect and mitigate memorization in diffusion models. The advantage of our proposed method is that it will not compromise the speed of either the training or the inference processes in these models while preserving the quality of generation. 
Our code is available at \href{https://github.com/renjie3/MemAttn}{github.com/renjie3/MemAttn}.
\end{abstract}

\input{secs/intro}
\input{secs/related}
\input{secs/preliminary}
\input{secs/property}
\input{secs/method}
\input{secs/experiments}

\section{Conclusion}
In this work, we present a novel view to understand the issue of memorization in diffusion models, focusing on its link with cross-attention mechanisms. We observe that cross attention often concentrates on the embeddings of trigger tokens, leading to memorization of specific training images. We further explore the inherent characteristics of cross attention and provide various insightful findings. Based on these findings, we introduce new methods to detect and mitigate the memorization without affecting the quality of the generated images and the models' speed during both training and inference. Experimental results support our findings and demonstrate the effectiveness of our proposed methods.

\section*{Acknowledgement}

Jie Ren, Han Xu, Yaxin Li, Shenglai Zeng, and Jiliang Tang are supported by the National Science Foundation (NSF) under grant numbers CNS 2246050, IIS1845081, IIS2212032, IIS2212144, IOS2107215, DUE 2234015, DRL 2025244 and IOS2035472, the Army Research Office (ARO) under grant number W911NF-21-1-0198, the Home Depot, Cisco Systems Inc, Amazon Faculty Award, Johnson\&Johnson, JP Morgan Faculty Award and SNAP.


%
%
\bibliographystyle{splncs04}
\bibliography{main}
\input{secs/append}
\end{document}

%% file: secs/intro.tex
\section{Introduction}
\label{sec:intro}

Recent advancements in diffusion models have demonstrated exceptional capabilities in image generation. Notably, text-to-image (T2I) diffusion models~\cite{ramesh2022hierarchical, rombach2022high} excel at producing high-quality images that adhere precisely to textual prompts. However, these models can memorize training data, including images and their relationship with the input textual prompts~\cite{carlini2023extracting, naseh2023understanding, naseh2023memory, webster2023reproducible}. Such a memorization issue poses a significant risk of copyright infringement of the training data and increases the possibility of privacy-sensitive information leakage~\cite{zhang2024counterfactual, jiang2023ai, carlini2022privacy}.

Existing investigations on this issue have primarily focused on identifying the data that is susceptible to causing memorization. For example, data duplication is found to lead to memorization, suggesting that de-duplication of training data could be a mitigation strategy~\cite{carlini2023extracting,somepalli2023diffusion, webster2023duplication}; and replicated captions are also revealed to contribute to this problem~\cite{somepalli2024understanding}. However, they largely overlook the understanding of how such data influences the behavior of the model. It is evident from a recent work~\cite{wen2023detecting} that the difference between the model output of a memorized prompt and an empty prompt is larger than that between a non-memorized prompt and an empty prompt. However, this finding still does not explain the inner behaviors of the model associated with memorization. 
%
    

\begin{figure}[tb]
  \centering
  \begin{subfigure}{0.24\linewidth}
  \centering
    \includegraphics[width=0.99\textwidth]{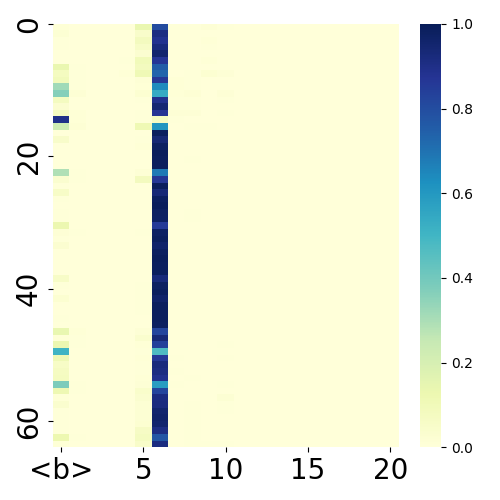}
    \captionsetup{justification=centering}
    \caption{Memorization \\ ($t=T$)}
    \label{fig:attn1}
  \end{subfigure}
  \begin{subfigure}{0.24\linewidth}
  \centering
    \includegraphics[width=0.99\textwidth]{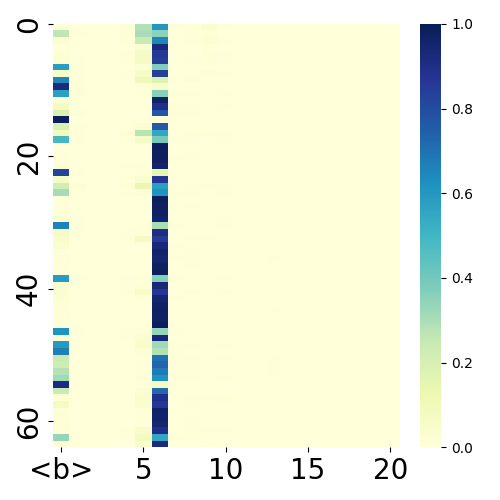}
    \captionsetup{justification=centering}
    \caption{Memorization \\ ($t=T/4$)}
    \label{fig:attn2}
  \end{subfigure}
  \begin{subfigure}{0.24\linewidth}
  \centering
    \includegraphics[width=0.99\textwidth]{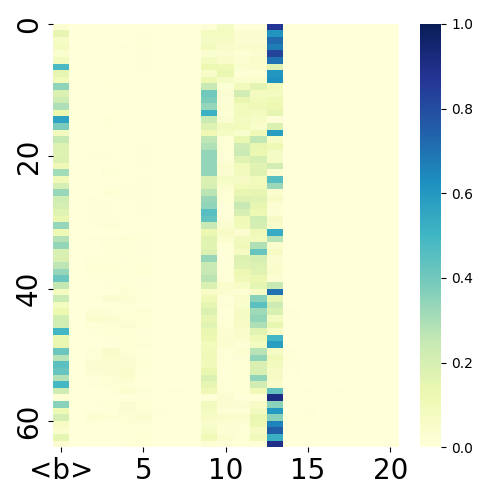}
    \captionsetup{justification=centering}
    \caption{Non-memorization \\ ($t=T$)}
    \label{fig:attn3}
  \end{subfigure}
  \begin{subfigure}{0.24\linewidth}
  \centering
    \includegraphics[width=0.99\textwidth]{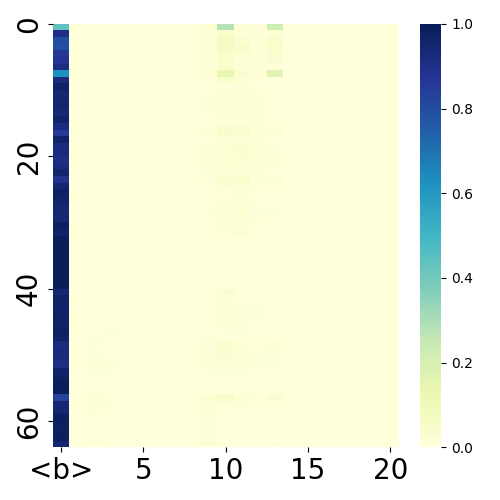}
    \captionsetup{justification=centering}
    \caption{Non-memorization \\ ($t=T/4$)}
    \label{fig:attn4}
  \end{subfigure}
  \caption{Cross-attention map for samples with and without memorization. ($X$ axis: token index; $Y$ axis: index of image representation dimension; $t$: diffusion generation step; $T$: total steps of diffusion process; <b>: token of <begin>.)}
  \label{fig:attn_map}
\end{figure}
%

In this work, we introduce a novel perspective to understand memorization via the behavior of \textbf{``cross-attention''}. Cross attention has been widely used by text-to-image diffusion models, serving as the primary mechanism of selecting information from the prompts to guide diffusion generation process\cite{rombach2022high, koh2021text, saharia2022photorealistic, ruiz2023dreambooth}.
Given that the memorized training images are usually triggered by the memorized textual prompts, cross attention could exhibit unique behaviors specific to memorization. Therefore, we investigate the relationship between the cross-attention and memorization. Specifically, we empirically observe a significant difference between the distributions of the cross-attention with/without memorization. One such example is demonstrated in \cref{fig:attn_map}. The memorized samples (\cref{fig:attn1} and \ref{fig:attn2}) tend to allocate most of the attention to the embeddings of specific tokens throughout all the diffusion steps. In contrast, non-memorization (\cref{fig:attn3} and \ref{fig:attn4}) has a more dispersed attention distribution. We deeply investigate the memorization behavior of cross attention and unveil insightful findings which pave us a way to design strategies to detect and mitigate memorization. Our detection approach is based on the quantification of the attention behavior in \cref{sec:property}. It only requires simple calculation on attention score, and does not need repeated or additional inference operations which are often required by existing methods~\cite{carlini2023extracting, wen2023detecting}.
Meanwhile, we propose an inference-time mitigation method and a training-time mitigation method that effectively reduce memorization by adjusting the attention dispersion. Similar to our detection approach, the proposed mitigation methods can preserve output quality and have little effect on image generation speed. 
Extensive experiments are conducted to validate our insights and the efficacy of our detection and mitigation strategies.



%% file: secs/related.tex
\section{Related works}
\subsubsection{Memorization.} Researchers have found that generative models tend to memorize and reproduce their training data~\cite{carlini2021extracting, carlini2023extracting, carlini2022quantifying, mireshghallah2022memorization, naseh2023understanding, naseh2023memory}, which brings significant privacy and copyright concerns~\cite{zhang2024counterfactual, jiang2023ai, carlini2022privacy}. Somepalli \etal~\cite{somepalli2023diffusion, somepalli2024understanding} show that diffusion models exhibit the behavior of producing verbatim training images. Based on this, Carlini \etal~\cite{carlini2023extracting} develop a strategy to extract data from diffusion models by generating numerous images to assess generation density. To address these concerns, researchers have explored several strategies, such as data de-duplication~\cite{carlini2023extracting, webster2023duplication} and data augmentations~\cite{daras2024ambient, somepalli2024understanding}. More recently, Wen \etal~\cite{wen2023detecting} propose a new mitigation based on their observation that the model output of memorized prompts has a significant difference from non-memorized prompt. 


\subsubsection{Cross Attention.}
Cross-attention mechanisms are instrumental in the T2I diffusion models~\cite{rombach2022high}, enabling the precise generation that correspond to textual prompts~\cite{koh2021text, saharia2022photorealistic, ruiz2023dreambooth}. 
Controlled T2I generation by modifying the cross attention has been widely proposed~\cite{wei2023elite, liu2023video, ma2023directed}. For example, 
cross-attention can be used to refine image editing~\cite{hertz2022prompt} and address issues like catastrophic neglect~\cite{chefer2023attend}. 
Besides, cross attention is also applied to text-image matching~\cite{lee2018stacked, wei2020multi} and explanation of T2I generation~\cite{tang2022daam}.
These works show the versatility and importance of cross attention in creating semantically coherent generation models. 

%% file: secs/preliminary.tex
\section{Preliminaries}

This section presents some essential preliminaries, including the generation process of diffusion models, the architecture of cross-attention connecting the prompt with generation, key definitions in prompts and our preliminary observations.

\subsection{Generation Process of Diffusion Models} 

Denoising Diffusion Probabilistic Models (DDPM)~\cite{ho2020denoising} typically involves a \textit{forward} diffusion process and a \textit{reverse} diffusion process. The forward process is a $T$-step Markov chain which transforms a data point $x_0$ from the target image distribution to a random Gaussian distribution. It introduces a small Gaussian noise at each step into the data point following
\begin{align}
    q\left(x_t \mid x_{t-1}\right)=\mathcal{N}(x_t ; \sqrt{1-\beta_t} x_{t-1}, \beta_t \mathbf{I}),
\end{align}
where $\beta_t$ is the variance schedule. The reverse process generates images by
\begin{align}\label{eqn:p}
    p_\theta\left(x_{t-1} \mid x_t\right)=\mathcal{N}\left(x_{t-1} ; \mu_\theta\left(x_t, t\right), \Sigma_\theta\left(x_t, t\right)\right),
\end{align}
where each reverse step follows a Gaussian distribution with $\mu_\theta$ and $\Sigma_\theta$ as some mean and variance. With a parameterized denoising network $\epsilon_\theta$, following~\cite{ho2020denoising},  the generation process can be explicitly expressed as
\begin{align}\label{eqn:formula}
    {x}_{t-1}=\frac{1}{\sqrt{\alpha_t}}\left({x}_t-\frac{1-\alpha_t}{\sqrt{1-\bar{\alpha}_t}} \epsilon_\theta\left({x}_t, t\right)\right)+\sigma_t w,
\end{align}
where $\alpha_t = 1-\beta_t$, $\bar{\alpha}_t=\prod_{i=1}^t\alpha_t$, $\sigma_t$ can be $\sqrt{\beta_t}$ or $\sqrt{\frac{1-\bar{\alpha}_{t-1}}{1-\bar{\alpha}_t} \beta_t}$, and $w \sim \mathcal{N}(\mathbf{0}, \mathbf{I})$. To connect Eq. (\ref{eqn:p}) with (\ref{eqn:formula}), $\mu_\theta(x_t,t)=(x_t-\frac{\beta_t}{\sqrt{1-\bar\alpha_t}}\epsilon_\theta(x_t,t))/\sqrt{\alpha_t}$, and $\Sigma_\theta(x_t,t)$ is determined by $\sigma_t$. Note that when generating new images in reverse process, we start from a random noise $x_T$ and end at step $0$ with the output as $x_0$.

\subsection{Cross Attention in Text-to-image Generation}
To guide the image generation process using extra conditions, Rombach \etal~\cite{rombach2022high} further propose Latent Diffusion Models (LDMs). Two changes are applied in LDMs compared to a vanilla diffusion model.
First, instead of directly generating images, the diffusion process proceeds with a low-dimensional latent representation of the image via the compression using a variational autoencoder (VAE)~\cite{kingma2013auto}. In this case, $I_0$ is the image, $\mathcal{E}$ is the image encoder of VAE, and we feed the latent representation $x_0 = \mathcal{E}(I_0)$ in diffusion models. Second, to introduce an extra condition $c$, $c$ works as an input of $\epsilon_\theta$ to guide the output of $\epsilon_\theta\left({x}_t, t, c\right)$. 


Extending from the basic framework of LDMs, Stable Diffusion (SD) takes textual prompts as condition $c$ to generate image described by given texts. SD first uses the text encoder of CLIP~\cite{radford2021learning} to get the prompt's embeddings, $e_c$, then uses the cross-attention mechanism to select the information and feed it into each of the hidden layers, $z_t$, in the U-Net backbone of diffusion models. In the cross-attention architecture, $Q(z_t)$, $K(e_c)$ and $V(e_c)$ are the three components connecting the prompt with the image. The terms $Q(z_t)$ and $K(e_c)$ decide the attention score allocated on each token embedding by 
\begin{align}
    A = \operatorname{softmax}\left(\frac{Q K^T}{\sqrt{d}}\right).
\end{align}
The term $V(e_c)$ is the textual information to be selected, and is further multiplied with $A$ in the cross attention and passed to the U-Net of the diffusion model. Besides, instead of a single-head cross attention, one may create multiple groups of $(Q,K,V)$ and merge them to a multi-head cross attention.  

When training the diffusion model, the training data consists of images and their paired textual prompts. In addition to the U-Net in the diffusion model, the cross attention is also trained to match the textual input and the generated image. The only components fixed in the training are the image encoder (VAE) and the text encoder, using which we can obtain the image latent representation, $x_0$, and the prompt embeddings, $e_c$. 

In this paper, we focus on the attention score $A$ to understand and mitigate the memorization. Each row of $A$ is a set of attention score for a hidden dimension of $z_t$ which assigns weights on the embeddings of different tokens of $V(e_c)$. Then the weighted information is aggregated to guide the generation process. Thus, the sum of attention on each row, \ie for each hidden dimension of $z_t$, is 100\%. For multi-head attention, each head calculates its corresponding attention score matrix respectively, and the sum of attention in each row of each head is 100\%. 

\subsection{Category of Tokens in the Prompts}
\label{sec:cate_token}
To facilitate the analysis of the connection between cross attention and memorization, we divide the tokens of textual prompts into three categories: \textbf{beginning} token, \textbf{prompt} tokens, and \textbf{summary} tokens. For example, in the prompt, ``<begin> \textit{A picture of mountains covered by snow} <end> <padding> <padding> ... <padding>'', the text encoder adds <begin> before the prompt tokens, and <end> <padding> after the prompt tokens. Given that the text encoder operates causally, the beginning token does not hold any semantic information derived from the textual prompt, as it exists before the entirety of the prompt.
In contrast, for <end> and <padding>, we refer them as ``\textit{summary} tokens'' since
the embeddings of these tokens encapsulate the semantic information of all preceding tokens.  

\subsection{Beginning Token Attains High Attention in Non-memorization}
\label{sec:begin_token}

From the attention maps of \cref{fig:attn_map} in \cref{sec:intro}, we find that the beginning token shows different patterns for samples with and without memorization. In non-memorization, although the attention distribution is disperse, the beginning token holds a high attention score, especially in \cref{fig:attn4}. Instead, in memorization, the beginning token is assigned with very little attention. To facilitate the understanding of memorization in the following sections, we first demonstrate the pattern of the beginning token's attention score in this subsection. We demonstrate the averaged attention score of the beginning token through all the diffusion steps in \cref{fig:token_0}. It can be seen that the beginning token has an attention score higher than 80\% in all the diffusion steps. Since the beginning token contains no semantic information, this implies that a small portion of the attention score is sufficient to collect the semantics from other tokens, especially in the late steps with a small $t$. 
Besides, in \cref{fig:token_0}, the attention of beginning tokens increases from step $T$ to step $0$. 
Intuitively, after the main body of the image is generated with a large $t$, the diffusion model mainly focuses on denoising when $t$ is small, which does not need much textual information from prompts.

%% file: secs/property.tex
\section{Cross Attention in Memorization}
\label{sec:property}


Cross attention plays an important role in text-to-image diffusion models, where it integrates the guidance of textual conditions into the image generation process. This mechanism facilitates that different segments of the generated image can pay attention to different tokens in the prompt and combine them reasonably in a global scope. In this section, we present our findings on memorization via cross attention through empirical studies. 
The experimental studies conducted in this section are based on SD v1.4 with the memorized prompts extracted by Webster~\cite{webster2023reproducible} and 500 non-memorized prompts generated by ChatGPT-4~\cite{achiam2023gpt}. In Appendix A and 
\cref{exp:detect}, we further show that the findings can also be generalized to SD v2.0, transformer-based diffusion model~\cite{chen2023pixart}, T5-based diffusion models~\cite{deepif, chen2023pixart} and fine-tuned SD~\cite{epiCRealism, wallace2023diffusion}.



\begin{figure}[tb]
    \centering
    \begin{minipage}{0.45\textwidth}
        \centering
        \includegraphics[width=0.84\textwidth]{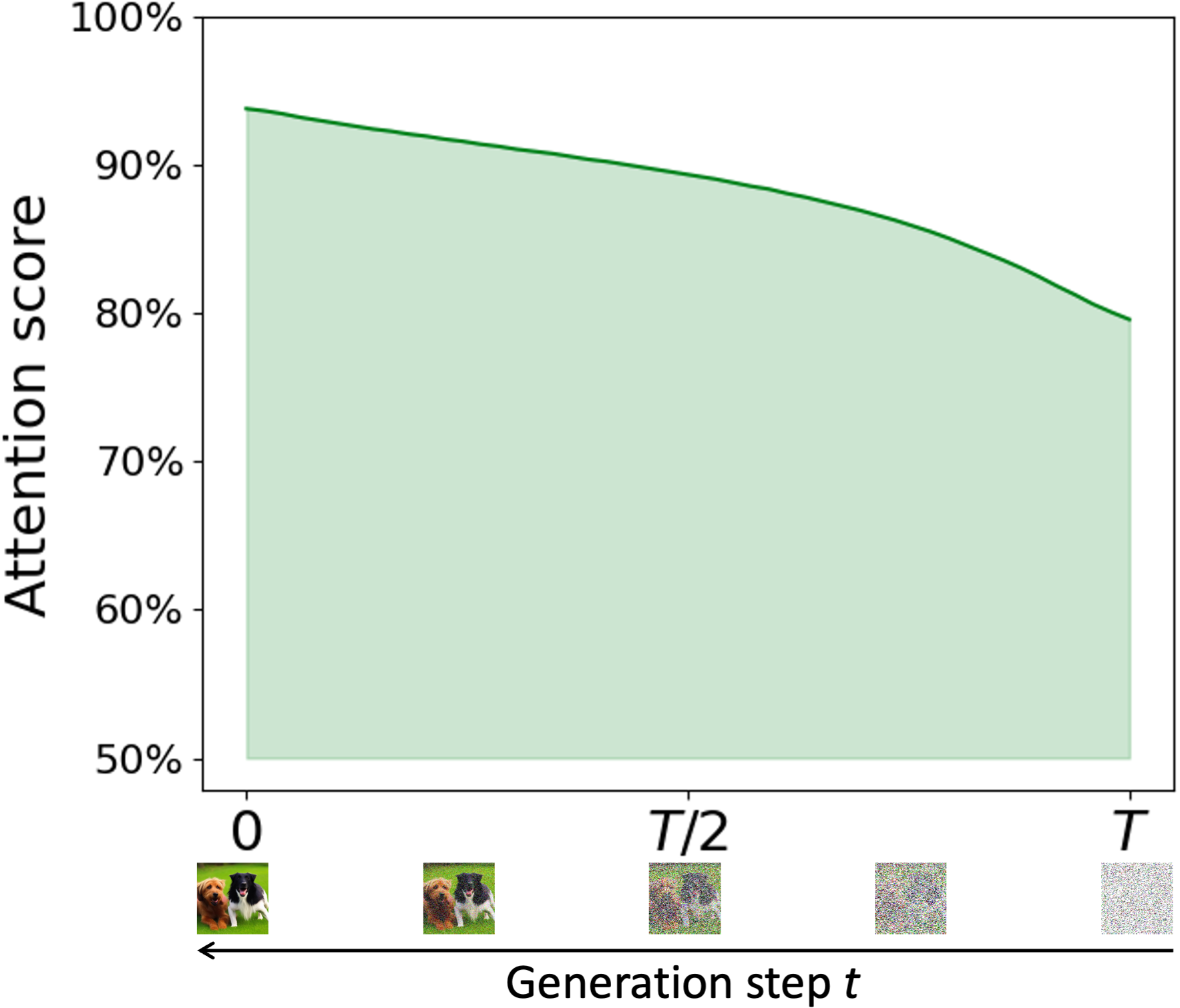} 
        \caption{Attention of beginning token}
        \label{fig:token_0}
    \end{minipage}\hfill
    \begin{minipage}{0.45\textwidth}
        \centering
        \includegraphics[width=0.84\textwidth]{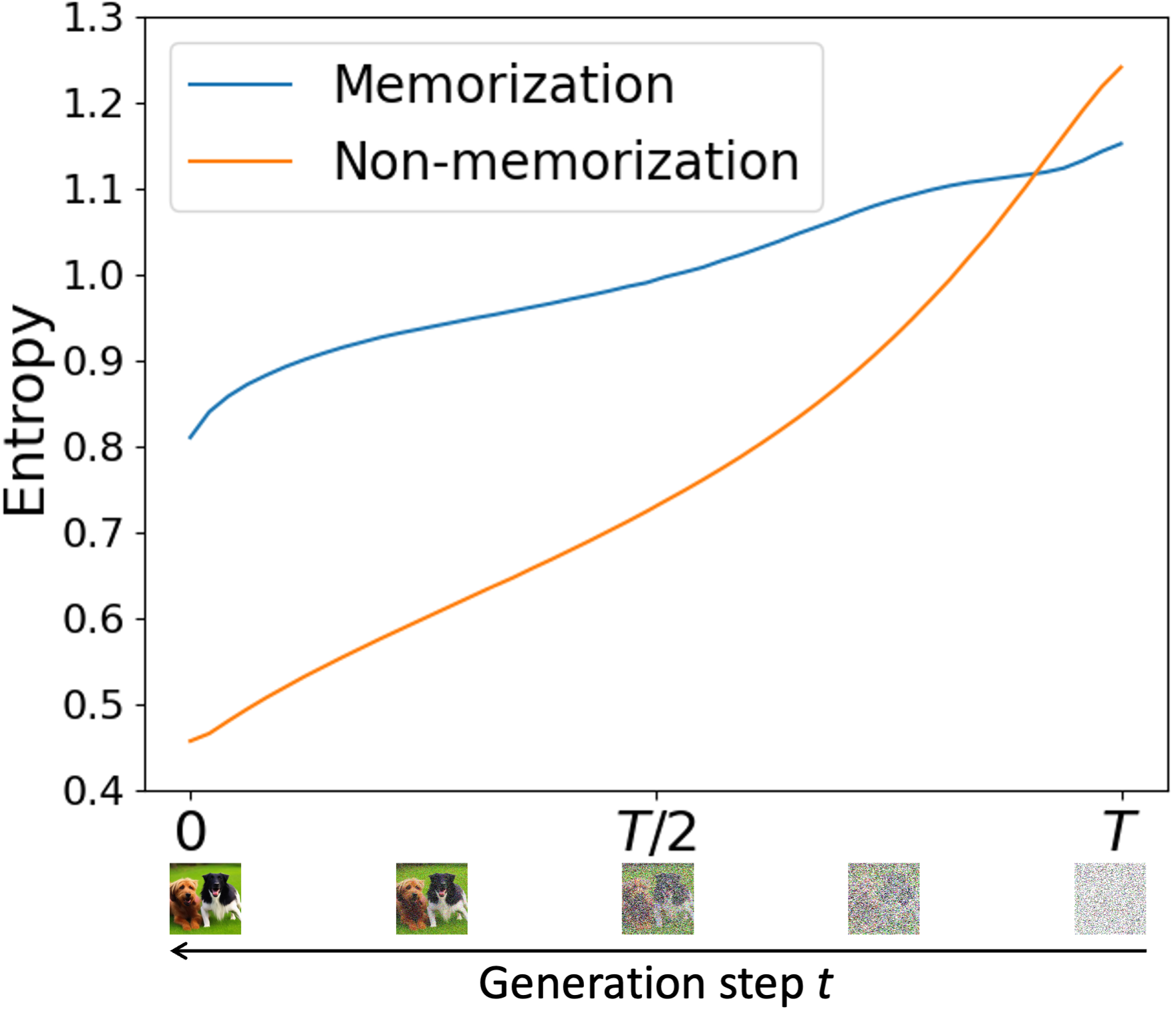} 
        \caption{Attention entropy of SD v1.4}
        \label{fig:entropy_sd1}
    \end{minipage}
\end{figure}


\subsection{Finding 1: Concentrated Attention Score on Trigger Tokens}
\label{sec:main_entropy}

\cref{fig:attn_map} in \cref{sec:intro} demonstrates that the memorized samples tend to allocate most of the attention to the embeddings of specific tokens. In this subsection, we further quantify this finding by defining attention entropy.

Recall that in cross attention, each hidden dimension in $z_t$ will allocate attention scores on the tokens. The sum of attention scores for each hidden dimension in each attention head is 100\%. Thus, we can consider it as a discrete probabilistic distribution. Meanwhile, entropy is usually used to measure the uncertainty and dispersion of probabilistic distributions. Thus, the entropy of attention can measure the dispersion of attention~\cite{pmlr-v202-zhai23a, vig2019analyzing}. It is defined as follows
\begin{align}
\label{eq:entropy}
    E_t =\sum_{i=1}^N-\overline{a}_i \log \left(\overline{a}_i\right),
\end{align}
where $N$ is the number of tokens, $t$ is diffusion step, and $\overline{a}_i$ is the average attention score on the $i$-th token. Unless otherwise stated, $\overline{a}_i$ is averaged across all U-Net layers, attention heads and hidden dimensions. Higher entropy indicates more disperse attention distribution. 

In~\cref{sec:begin_token}, we find that the cross-attention score of non-memorization will gradually concentrate on the beginning token. In contrast, as shown in \cref{fig:attn_map}, for the prompts of memorization, more attention will be concentrated on the embeddings of specific prompt tokens and summary tokens throughout all the steps in the generation process, referred to as \textit{trigger} tokens. Meanwhile, the memorized samples will distract from the beginning token. This leads to a disperse attention distribution. Therefore, we expect that the attention entropy is low for non-memorization samples, and high for memorized ones. 



We demonstrate the entropy of different diffusion steps in \cref{fig:entropy_sd1} to verify our intuitions. In non-memorization samples, the entropy reduces fast from step $T$ to step $0$, which is consistent with the fact that the attention is gradually concentrated into the beginning token.
On the other hand, for memorized samples, the entropy becomes higher than non-memorization especially when $t$ is small, which also aligns with our observation that the model pays high averaged attention scores on trigger tokens. One exception is that, when $t$ is large, \eg $t=T$ in \cref{fig:attn3}, compared to the memorization case, the non-memorization prompts also allocate higher attentions on prompt and summary tokens, resulting in a higher entropy in \cref{fig:entropy_sd1}. To understand this, although non-memorization in general imposes the highest weight on the meaningless beginning token, to generate proper images, it still needs to correctly understand the input prompt. As a result, for large $t$, the model focuses on collecting semantic information from different tokens, and the corresponding entropy is high. In contrast, with memorization, the model only needs to focus on the trigger tokens, and it is unnecessary to further collect information from other tokens, resulting in a lower entropy for larger $t$. 

With the quantification of the attention score via entropy, we verify our finding about the disproportionate attention on \textit{trigger} tokens.


\subsection{Finding 2: Different Types of Memorization Focus on Different Types of Tokens}
\label{sec:diff_type}

Existing studies suggest that there exist different types of memorized prompts in diffusion models~\cite{webster2023reproducible}. In this subsection, we discuss cross attention behaviors {on} different types of memorized prompts, and provide a deeper understanding about the connection between cross attention and memorization.

In literature, Webster~\cite{webster2023reproducible} provides the dataset of memorized prompts and divides them into three types: 
\begin{itemize}
    \item Matching memorization (MM): one memorized prompt generates one image that is exactly matching with the original paired training image. 
    \item Retrieval memorization (RM): one memorized prompt generates images that match with a subset of training images.
    \item Template memorization (TM): a group of prompts generate images highly align with training images, but may have variations in colors or styles. 
\end{itemize}

\begin{figure}[tb]
  \centering
  \begin{subfigure}{0.32\linewidth}
  \centering
    \includegraphics[width=0.90\textwidth]{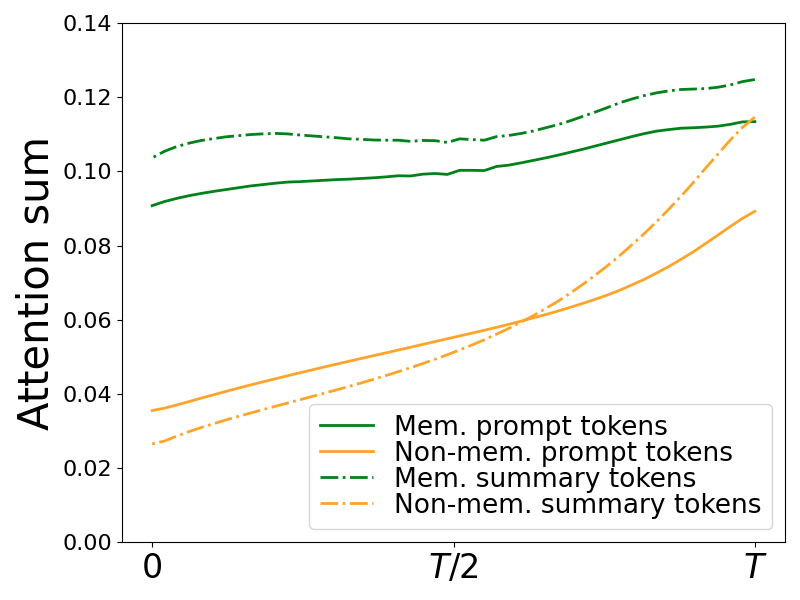}
    \caption{MM}
    \label{fig:attn_sum_mv}
  \end{subfigure}
  \begin{subfigure}{0.32\linewidth}
  \centering
    \includegraphics[width=0.90\textwidth]{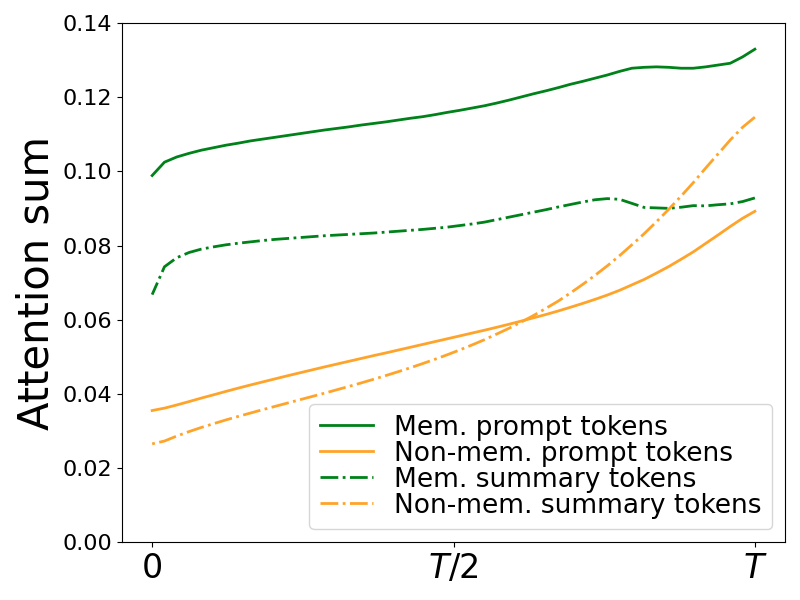}
    \caption{RM}
    \label{fig:attn_sum_rv}
  \end{subfigure}
  \begin{subfigure}{0.32\linewidth}
  \centering
    \includegraphics[width=0.90\textwidth]{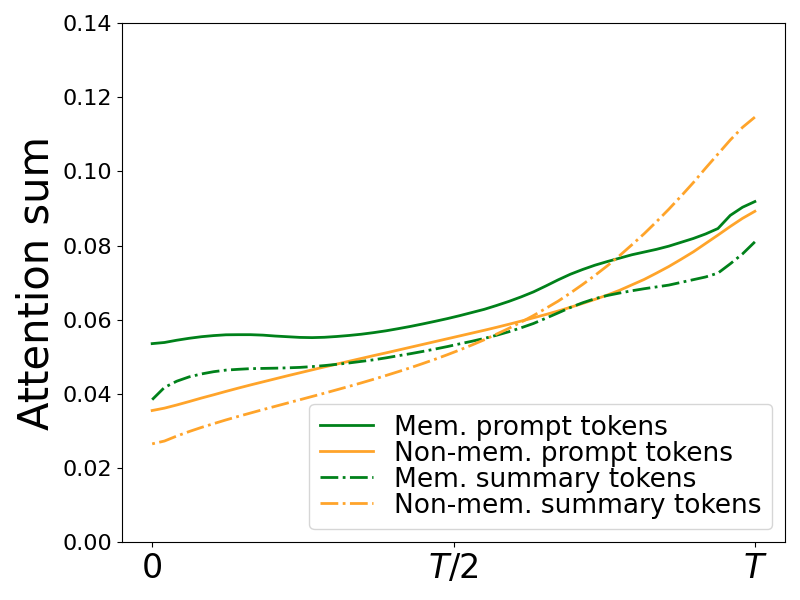}
    \caption{TM}
    \label{fig:attn_sum_tv}
  \end{subfigure}

  \begin{subfigure}{0.32\linewidth}
  \centering
    \includegraphics[width=0.99\textwidth]{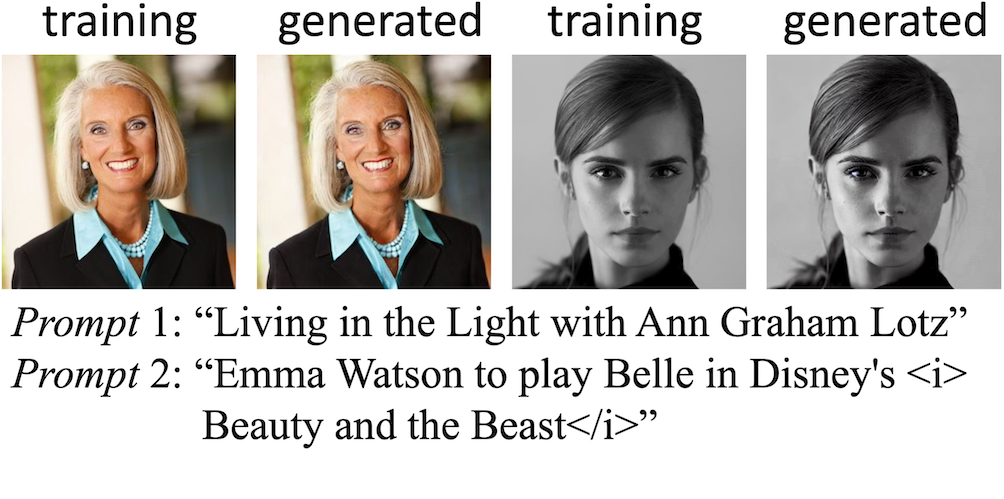}
    \caption{MM}
    \label{fig:ex_mm}
  \end{subfigure}
  \begin{subfigure}{0.32\linewidth}
  \centering
    \includegraphics[width=0.99\textwidth]{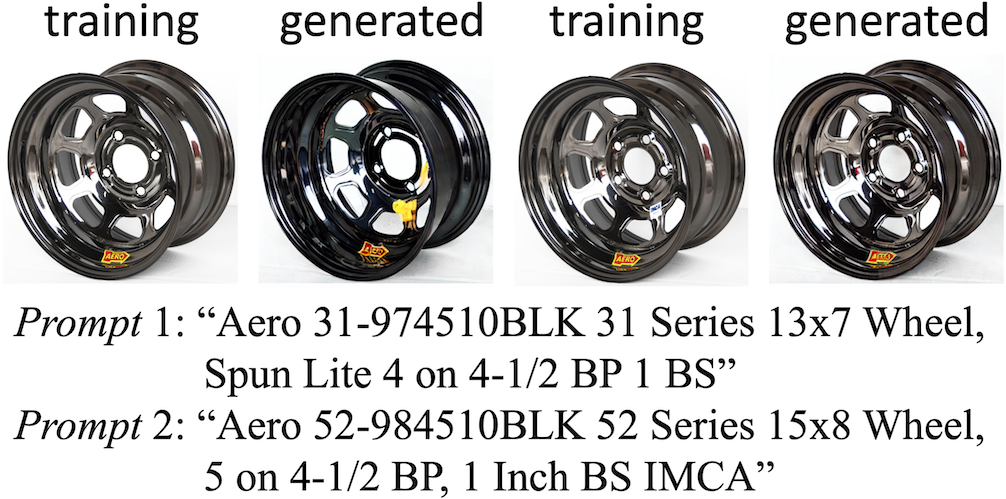}
    \caption{RM}
    \label{fig:ex_rm}
  \end{subfigure}
  \begin{subfigure}{0.32\linewidth}
  \centering
    \includegraphics[width=0.99\textwidth]{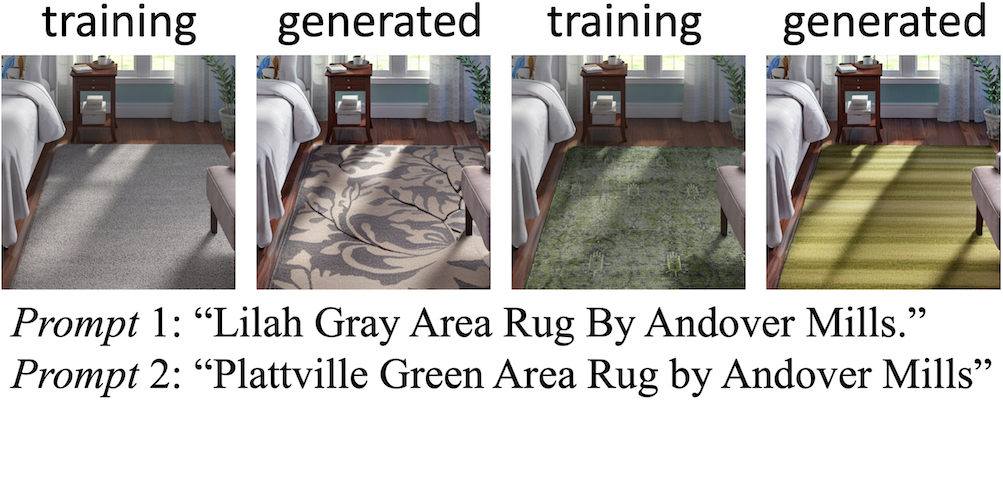}
    \caption{TM}
    \label{fig:ex_tm}
  \end{subfigure}
  \caption{(a)(b)(c) Sum of attention score on prompt tokens and summary tokens in MM, RM and TM; (d)(e)(f) Examples of prompts of MM, RM and TM (In each sub-figure, the left pair of training and generated images is associated with \textit{Prompt} 1, and the right pair is associated with \textit{Prompt} 2).}
  \label{fig:three_types}
\end{figure}

We demonstrate the sum of attention score on prompt and summary tokens respectively in \cref{fig:three_types} to examine to what extent the model focuses on these two types of tokens, and also give prompt examples of each types in \cref{fig:three_types}. We make the following two observations from \cref{fig:three_types}.

First, MM pays more attention to summary tokens compared with RM and TM. In MM, the memorized prompts are totally different from each other. Since the summary tokens contain the semantics of the whole sentence, overfitting to the summary tokens can memorize the unique prompts and connect them with the paired training images easier. In contrast, for RM and TM, the memorized prompts share the same sentence template and a few identical tokens in a group of prompts. The model overfits these shared tokens and overlooks the difference. 

Second, all three types of memorization have a significantly slower reduction of summary-token attention score than non-memorization from step $T$ to step $0$. It implies that non-memorization distracts from summary tokens faster. We conjecture that this is because different parts of the non-memorized generated images focus on different semantics of the prompt rather than the summary of the prompt. For example, in the prompt ``\textit{Two dogs playing on the grass}'', some pixels focus on ``\textit{dogs}'', and some focus on ``\textit{grass}''. 
But for memorization, it does not distinguish prompt tokens or summary tokens like non-memorization since the overfitting of trigger tokens can simply guide each pixel in the image.
This characteristic can distinguish between memorization and non-memorization generations.

\subsection{Finding 3: Concentration is More Active in Certain U-Net Layers}
\label{sec:certain_layer}


According to \textbf{Finding 1} in \cref{sec:main_entropy}, the entropy is better distinguished between memorization and non-memorization in step $0$ than step $T$. This means the whole generation is involved since step $0$ is the last step of the generation. However, in this subsection, we find that different U-Net layers have different behaviors for memorization, and in some active U-Net layers, the memorization is already distinguished at the first step, \ie when $t=T$. 

\begin{figure}[tb]
  \centering
  \includegraphics[width=0.6\textwidth]{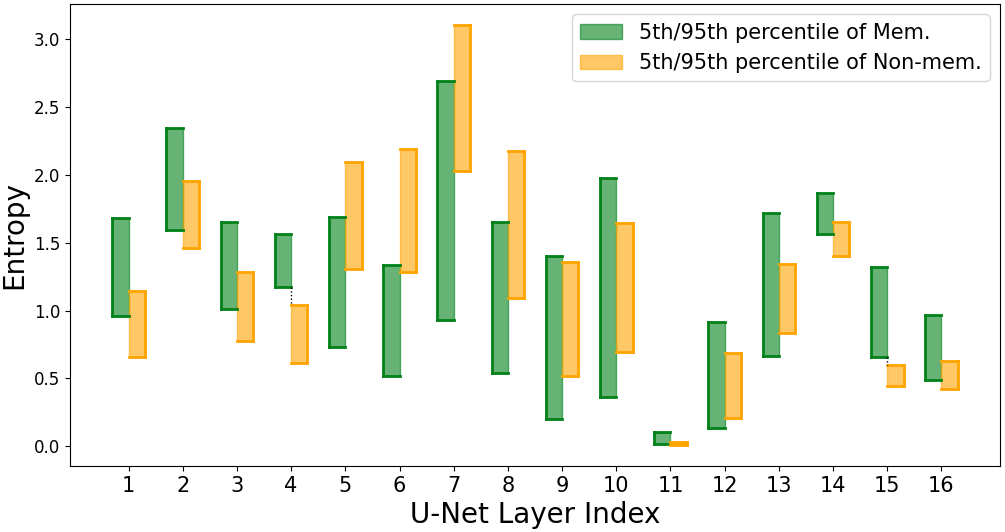}
  \caption{Entropy of each U-Net layer in SDv1.4}
  \label{fig:single_layer_95}
\end{figure}
In SD, there are 16 U-Net layers including down-sampling, middle and up-sampling layers. The cross-attention mechanism has different sets of multi-head $(Q, K, V)$ for each layer. These sets collect different semantic information and feed them into each layer.
We find that the concentration on special tokens is not uniform on all the layers. We show the entropy of each layer in the first diffusion step, \ie $t=T$, in \cref{fig:single_layer_95}. In the first diffusion step, the entropy of distinct U-Net layers have different overlapping between memorization and non-memorization. The fourth and fifteenth layers have clearer separation, which can distinguish memorization and non-memorization better. This shows that using the first diffusion step has the potential to detect the memorized samples in one step.



%% file: secs/method.tex
\section{Detection and Mitigation}

Building upon the above findings, we propose methods to detect and mitigate memorization effect via cross attention. Our methods are superior to previous ones~\cite{carlini2023extracting, wen2023detecting} since we require almost no extra computation and will not compromise the speed of training or inference, while preserving the quality of generation. 

\subsection{Memorization Detection}
\label{sec:detect}


\textbf{Finding 1} to \textbf{3} in \Cref{sec:property} demonstrate different attention distributions between memorization and non-memorization. In this subsection, we introduce two metrics guided by our findings to detect the memorization. 

The first metric is designed based on \textbf{Finding 1} and \textbf{Finding 2} as:
\begin{align}
    D=\frac{1}{T_D}\sum_{t=0}^{T_D - 1} E_t - \frac{1}{T_D}\sum_{t=0}^{T_D - 1} \left| E_t^{\text{summary}} - E_T^{\text{summary}} \right|,
    \label{eq:measure1}
\end{align}
where $E_t^{\text{summary}} = \sum_{i=N-S+1}^N-\bar{a}_i \log \left(\bar{a}_i\right)$ is the entropy calculated only on the summary tokens and $S$ is the number of summary tokens. The term $T_D$ is the number of generation steps used in calculating $D$. For SD, we choose $T_D = \frac{T}{5}$, which means we use the last $\frac{T}{5}$ steps of the reverse diffusion process to calculate $D$. The first term is inspired by \textbf{Finding 1} that trigger tokens will cause higher entropy especially than non-memorization when $t$ is close to $0$. The second term is from \textbf{Finding 2} that the sum of attention score of summary tokens in memorization will reduce slower than non-memorization. In \cref{eq:measure1}, we use $E_t^{\text{summary}}$ to replace sum to make its numerical values comparable with the first term, otherwise it becomes a negligible term. In Appendix B, we show that the sum and $E_t^{\text{summary}}$ have consistent conclusion with \textbf{Finding 2}. 

The second metric is motivated by \textbf{Finding 3}, which is defined as
\begin{align}
    E^l_{t=T}=\sum_{i=1}^N-\bar{a}_i^l \log \left(\bar{a}_i^l\right),
\end{align}
where $\bar{a}_i^l$ is the averaged attention on $l$-th layer of U-Net of the first step in the reverse diffusion process, \ie $t=T$. The major difference between the two metrics is that the second metric only requires the first step of diffusion process. Although the second requires an additional hyper-parameter $l$, it allows the model builders to detect the memorization instantly, which can save time and computation before adapting the follow-up process like mitigation. 




\subsection{Inference-time Mitigation}

In this subsection, we propose a method to mitigate memorization at the inference stage by reducing the weight on trigger tokens. This method consists of masking out the summary tokens and increasing the logits of the beginning token.

First, from \textbf{Finding 2}, we know that memorization is highly related to the summary tokens, while attention score on summary tokens drops quickly in non-memorization. Consequently, to control memorization, we mask out these summary tokens. 

Second, based on \textbf{Finding 1}, trigger tokens tend to have the large attention scores except for the beginning token. As a result, we reversely enlarge the attention score for the beginning token via adjusting the input logits of the softmax operator in the cross attention. To be specific, denote the original input logits as $\boldsymbol{s} = (s_1, s_2, ..., s_N)$ where $s_i$ is the logit of the $i$-th token, the re-scaled logit vector $\boldsymbol{s}^{\prime}$ is 
\begin{align}
    \boldsymbol{s}^{\prime} = (Cs_1, s_2, ..., s_N).
    \label{eq:rescale}
\end{align}
where $C$ is a factor to be applied on the beginning token. 
When $C > 1$, the attention score of beginning token is increased and the attention scores of other tokens become smaller. 

While our primary interest on $C>1$ is to enlarge the attention score of the beginning token, we also note that the re-scaling operation in \cref{eq:rescale} will also give a larger reduction on the tokens with larger attention scores. To explain this, when taking the gradient of each attention score w.r.t. $C$, we obtain
\begin{align}
    \frac{\partial \text{softmax} (\boldsymbol{s})_i}{\partial C} = - \frac{s_1 e^{Cs_1}}{(e^{Cs_1} + \sum_{j=2}^Ne^{s_j})^2} e^{s_i}
    \label{eq:grad_rescale}.
\end{align}
Since ${s_1 e^{Cs_1}}/{(e^{Cs_1} + \sum_{j=2}^Ne^{s_j})^2}$ is the same for $\forall i \in [2,N]$, the reduction ratio depends on $- e^{s_i}$, which means token with larger attention will be reduced more.

By combining the above two strategies, our final adjusted logits become
\begin{align}
    \boldsymbol{s}^{\prime} = (Cs_1, s_2, ..., s_{N-S}, -\infty, ..., -\infty).
    \label{eq:final_rescale}
\end{align}
This method only requires applying a mask and a re-scaling factor to the logits, and does not include any extra computation. Nonetheless, it can effectively mitigate memorization during inference and has little impact on generation quality.

\textit{Remark.} In the experiments in \cref{fig:infer_miti}, we show that, although detection has almost no extra computation, it is not necessary to detect before mitigation since the mitigation does not compromise the generation quality for both memorization and non-memorization. 
\subsection{Training-time Mitigation}

During training, we can mitigate memorization by removing the samples whose attention entropy is higher than a pre-defined threshold from the mini-batch. This idea is similar to~\cite{wen2023detecting}, in which it filters out the samples if their outputs of the diffusion model are significantly different from these of the empty prompt. However, the method in \cite{wen2023detecting} uses additional inference operation of the empty prompt, which requires more computation. In contrast, our approach eliminates the need for extra inference. We only compute the attention entropy as defined by \cref{eq:entropy} and remove the samples with high-entropy, which incurs a negligible computational cost. This alone suffices to effectively reduce memorization.


    

%% file: secs/experiments.tex
\section{Experiment}
\label{sec:exp}

In this section, we provide experiments to further support  \textbf{Finding 1} and demonstrate the effectiveness of the proposed methods. In \cref{exp:detect} and \cref{exp:mitigation}, we evaluate our performance on detection and mitigation, respectively. 
In \cref{exp:ablation}, we conduct ablation study on the two components of inference-time mitigation. 
Due to space limitation, additional experiments and details including a further verification showing the memorization's reliance on the embeddings of trigger tokens, a validation of our findings with SD v2.0, and selection of the pre-defined threshold for training-time mitigation are elaborated in Appendix A and C.

\subsection{Experiment Settings}

\textbf{Diffusion models and datasets.} We conduct the experiments using SD v1.4 and SD v2.0. We use the dataset of memorized prompts extracted by Webster~\cite{webster2023reproducible} as memorized samples, and use 500 prompts generated by ChatGPT-4~\cite{achiam2023gpt} as non-memorized samples. For the fine-tuning data in training-time mitigation, since duplicated data can cause memorization, we use 200 text-image pairs duplicated 50 times as memorized data and 20,000 captioned images from LAION~\cite{schuhmann2022laion} as non-memorized data, following the procedure of \cite{wen2023detecting}. 

\textbf{Baselines.} For \textit{detection}, we use two baselines for comparison. The first is the method from Carlini \etal~\cite{carlini2023extracting}. They observe that the model generates the same image for the memorized prompt when using different random seeds. Thus they propose to detect memorization by the $l_2$ distance of the images generated using different $n$ seeds. The second is from Wen \etal~\cite{wen2023detecting}. {Their detection method is built upon the observation where the difference of the output between memorized prompt and empty prompt is larger than the difference between non-memorized and empty prompt. To stabilize the method, they generate the image for $n$ times. For \textit{mitigation}, we compare with a the method from Wen \etal~\cite{wen2023detecting}. More details can be found at Appendix C.1.

\textbf{Evaluation metrics.} The \textit{detection} metrics include the area under the receiver operating characteristic curve (AUROC), the true positive rate at the false positive rate of 3\% (TPR@3\%FPR) and the time cost. The \textit{mitigation} metrics include Similarity Score~\cite{pizzi2022self, somepalli2023understanding} that measures the degree of memorization by the similarity between generated images and original training images {(higher similarity means more severe memorization)}; Fréchet Inception Distance (FID)~\cite{heusel2017gans} that measures the generation quality in realism and diversity {(lower FID indicates better quality)}; CLIP score~\cite{radford2021learning} that measures fine-tuning performance {(higher CLIP score means better fine-tuning performance);} and the time cost. Details can be found at Appendix C.2.

\subsection{Memorization Detection}
\label{exp:detect}

\begin{table}[tb]
\centering
\caption{Results of detection in AUROC, TPR@3\%FPR and time in seconds. ($s$: the number of steps used for detection; $n$: the number of generations required per prompt.)}
\label{tab:detection}
\resizebox{0.91\textwidth}{!}{
\begin{tabular}{cccc}
\toprule
\multirow{2}{*}{~~ Method ~~ } & \multirow{2}{*}{~ $s, n$ ~ } & SD v1.4 & SD v2.0 \\
 & & ~ AUROC / TPR@3\%FPR / {Time} ~ & ~ AUROC / TPR@3\%FPR / Time ~ \\
\midrule
\multirow{4}{*}{\makecell{Carlini \\ \etal \cite{carlini2023extracting}}} 
   & 1, 32  & 0.4982 / 0.020 / 3.724  & 0.5433  / 0.034   / 8.799 \\
   & 50, 4 & 0.9357 / 0.716 / 7.006  & 0.5305 / 0.280  / 16.901 \\
   & 50, 16 & 0.9983 / 0.980  / 27.983 & 0.9304 / 0.428 / 67.565 \\
   & 50, 32  & 0.9977 / 0.976  / 56.016 & 0.9336 / 0.428 / 135.770 \\ 
\midrule
\multirow{6}{*}{\makecell{Wen \\ \etal \cite{wen2023detecting}}}  & 1, 1 & 0.9662 / 0.835 / 0.132 & 0.9215 / 0.580 / 0.315\\
   & 1, 4  & 0.9967 / 0.977 / 0.521 & 0.9610 / 0.790 / 1.273 \\
   & 1, 32  & 0.9992 / 0.994 / 4.133 & 0.9744 / 0.863 / 10.531 \\
   & 50, 1 & 0.9957  / 0.988 / 2.582  & 0.9831 / 0.963 / 6.072 \\
   & 50, 4 & 0.9968 / 0.994 / 10.339 & 0.9886 / 0.973 / 24.267 \\
   & 50, 32 & 0.9972 / \textbf{0.997} / 41.378 & 0.9890 / \textbf{0.982} / 97.336 \\
\midrule
Ours - $D$ ~~~ & 50, 1 & \textbf{0.9998} / \textbf{0.997} / 1.745 & 0.9783   / 0.881 / 4.220 \\ 
Ours - $E^{l=4}_{t=T}$ & 1, 1 & 0.9933 / 0.980 / \textbf{0.116} & \textbf{0.9968} / 0.977 / \textbf{0.281} \\
\bottomrule
\end{tabular}
}
\end{table}
In this subsection, we present the detection performance of the two proposed methods, $D$ and $E^{l}_{t=T}$, in both accuracy and effeciency with \cref{tab:detection}. For all the detection, we use $l=4$.


To compare our method $D$ with the benchmarks, we can see that it costs 1.745 seconds on SD v1.4 and 4.220 seconds on SD v2.0, which is more efficient than most of the methods except the two fast settings $(s=1, n=1)$ and $(s=1, n=4)$ from Wen \etal~\cite{wen2023detecting}. However, comparing to these fast settings of Wen \etal~\cite{wen2023detecting}, our AUROC and TPR are much better, especially on SD v1.4 with almost perfect AUROC of 0.9997 and TPR of 0.997. 

In terms of our method $E^{l=4}_{t=T}$, it is the most efficient one and achieves good performance with AUROC higher than 0.993 and TPR higher than 0.977 for both SD v1.4 and v2.0. On SD v2.0, our method $E^{l=4}_{t=T}$ is the best one in both AUROC and detection speed. It is even the only method that achieves an AUROC higher than 0.99 on SD v2.0. On SD v1.4, although some of the other methods, e.g., $(s=50, n=32)$ of Wen \etal~\cite{wen2023detecting}, are better in AUROC and TPR than $E^{l=4}_{t=T}$, their time cost is 20 to 400 times of the method $E^{l=4}_{t=T}$.

\subsection{Memorization Mitigation}
\label{exp:mitigation}

\begin{figure}[tb]
  \centering
  \begin{subfigure}{0.23\linewidth}
  \centering
    \includegraphics[width=0.99\textwidth]{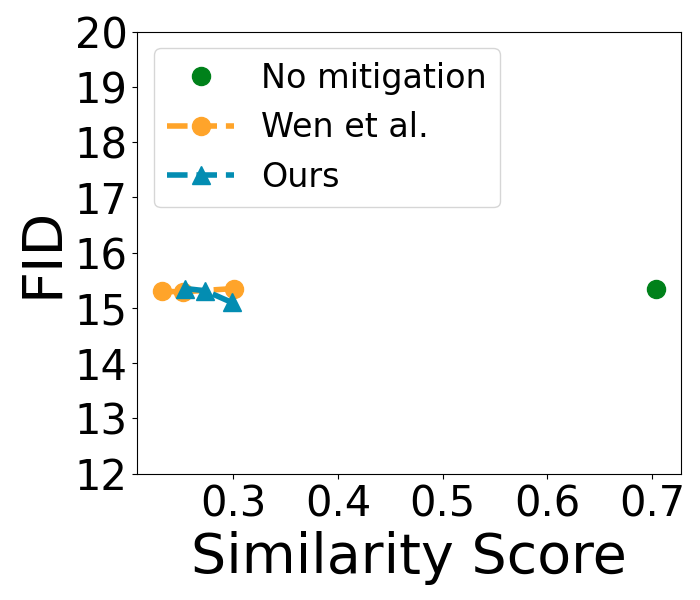}
    \caption{}
    \label{fig:infer_miti}
  \end{subfigure}
  \begin{subfigure}{0.23\linewidth}
  \centering
    \includegraphics[width=0.99\textwidth]{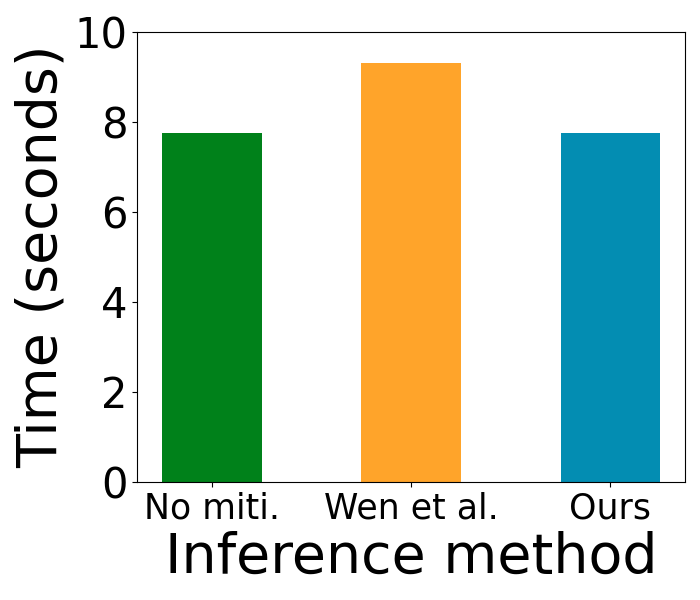}
    \caption{}
    \label{fig:infer_time}
  \end{subfigure}
  \begin{subfigure}{0.265\linewidth}
  \centering
    \includegraphics[width=0.99\textwidth]{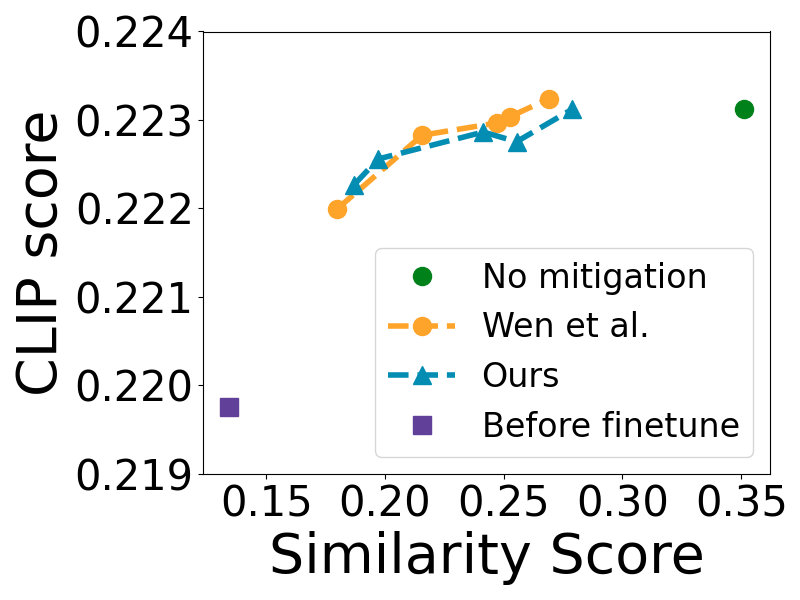}
    \caption{}
    \label{fig:train_miti}
  \end{subfigure}
  \begin{subfigure}{0.23\linewidth}
  \centering
    \includegraphics[width=0.99\textwidth]{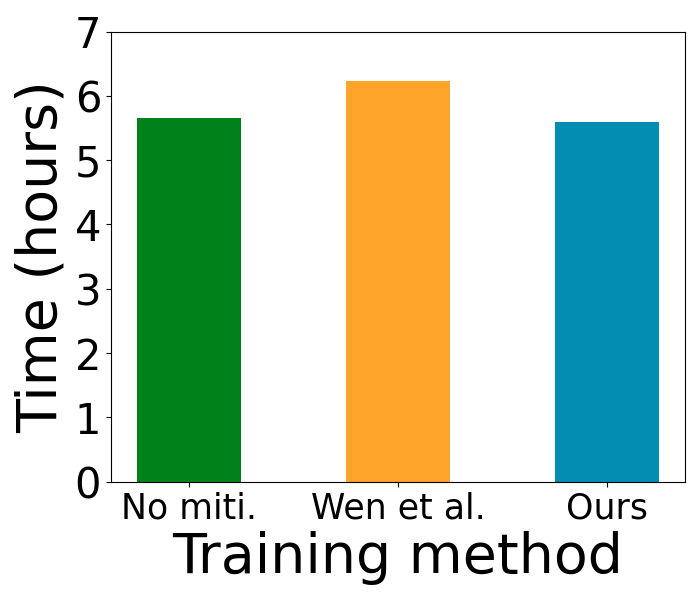}
    \caption{}
    \label{fig:train_time}
  \end{subfigure}
  \caption{(a)(b) Inference-time mitigation with similarity score, FID and running time. (c)(d) Training-time mitigation with similarity score, CLIP score and running time.}
  \label{fig:mitigation}
\end{figure}

In this subsection, we show that our mitigation methods can effectively reduce the memorization without sacrifice of the speed and the generation quality. 

For \textit{inference-time} mitigation, \cref{fig:infer_miti} presents the results of similarity score and FID with and without mitigation. The FID is calculated between generated images (including both memorized prompts from~\cite{webster2023reproducible} and 10,000 non-memorized prompts) and a subset of LAION~\cite{schuhmann2022laion} with 10,000 images. Similarity Score is calculated between the generated images by memorized prompts and the training images paired with the memorized prompts. When there is no mitigation, Similarity Score is 0.7. Our method can significantly reduce Similarity Scores from 0.7 to the range of 0.25 to 0.3 (by setting $C$ from 1.1 to 1.25). 
FID has almost no change after mitigation, which means the mitigation will not influence generation quality. Although the method of Wen \etal \cite{wen2023detecting} has similar mitigation effect, it is slower than ours by 20\% as shown in \cref{fig:infer_time}. Notably, our method maintains the same computational efficiency as performing inference without any mitigation.

For \textit{training-time} mitigation, we compare the mitigation effect and the fine-tuning performance with the baseline method from Wen \etal~\cite{wen2023detecting} in \cref{fig:train_miti}. The fine-tuning performance is measured by CLIP score which estimates the alignment between prompts and images. Higher CLIP score after fine-tuning indicates that the fine-tuned model can generate images that aligns the prompts better.
Both our method and the baseline effectively reduce similarity to around 0.2, with negligible impact on CLIP score. However, the method of Wen \etal~\cite{wen2023detecting} incurs a 10\% increase in training time. In contrast, our approach maintains the same training duration as scenarios without mitigation as shown in \cref{fig:train_time}.


\subsection{Ablation Study on Inference-Time Mitigation}
\label{exp:ablation}

\begin{figure}[tb]
  \centering
  \begin{subfigure}{0.08\linewidth}
  \centering
    \includegraphics[width=0.905\textwidth]{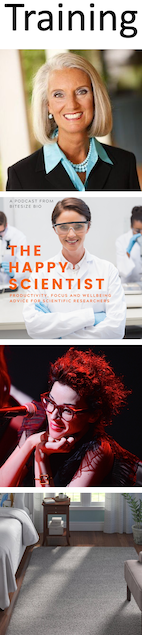}
    \captionsetup{margin={4pt,0pt}}
    \caption{}
    \label{fig:merged_image_training}
  \end{subfigure}
  \begin{subfigure}{0.16\linewidth}
  \centering
    \includegraphics[width=0.99\textwidth]{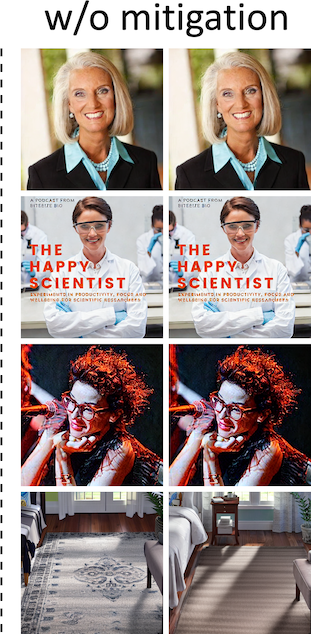}
    \captionsetup{margin={4pt,0pt}}
    \caption{}
    \label{fig:merged_image_no_miti}
  \end{subfigure}
  \begin{subfigure}{0.16\linewidth}
  \centering
    \includegraphics[width=0.99\textwidth]{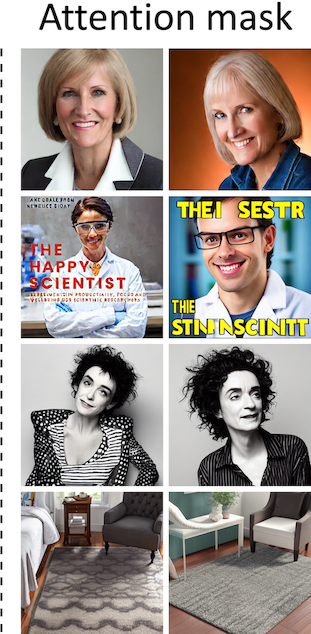}
    \captionsetup{margin={4pt,0pt}}
    \caption{}
    \label{fig:merged_image_attn_mask}
  \end{subfigure}
  \begin{subfigure}{0.16\linewidth}
  \centering
    \includegraphics[width=0.99\textwidth]{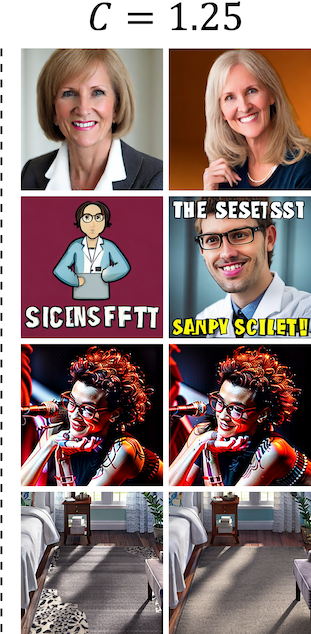}
    \captionsetup{margin={4pt,0pt}}
    \caption{}
    \label{fig:merged_image_miti1.25}
  \end{subfigure}
  \begin{subfigure}{0.16\linewidth}
  \centering
    \includegraphics[width=0.99\textwidth]{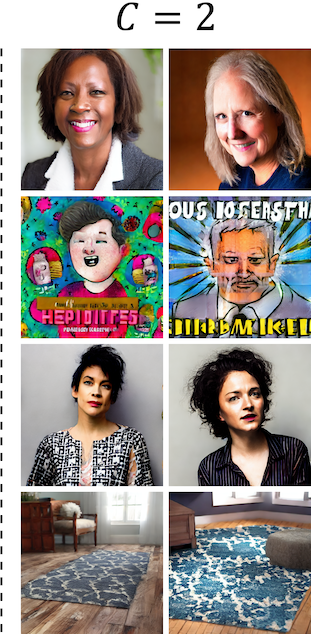}
    \captionsetup{margin={4pt,0pt}}
    \caption{}
    \label{fig:merged_image_miti2}
  \end{subfigure}
  \begin{subfigure}{0.16\linewidth}
  \centering
    \includegraphics[width=0.99\textwidth]{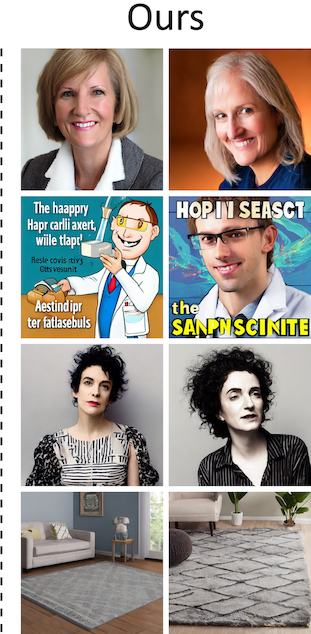}
    \captionsetup{margin={4pt,0pt}}
    \caption{}
    \label{fig:ours_infer_miti}
  \end{subfigure}
  \caption{Inference-time mitigation. Each row is generated with the same prompt. From (b) to (f), each prompt is generated twice with two seeds. In each row, the seeds for all the images on the left side are the same, and the seeds for the images on the right side are the same.}
  \label{fig:case}
\end{figure}
In this subsection, we conduct ablation study on the two components of inference-time mitigation, \ie logits re-scaling and summary token mask. We present a set of generated examples in \cref{fig:case}. 
Our method is shown in \cref{fig:ours_infer_miti} with summary tokens masked out and re-scaling factor $C=1.25$. It can effectively mitigate the memorization compared with no mitigation in \cref{fig:merged_image_no_miti}. However, if we only apply attention mask or logits re-scaling as shown in \cref{fig:merged_image_attn_mask} and \cref{fig:merged_image_miti1.25}, some memorized samples cannot be eliminated. If we increase $C$ to 2, although the memorization can be prevented, it might cause the loss of some semantic information like the identities and the appearance of the specific celebrities in \cref{fig:merged_image_miti2}. Due to the difference of formulations between \textbf{Finding 1} and \textbf{Finding 2}, the two components of our inference-time mitigation method may handle different memorized samples as shown above. On the other hand, when applying them together, they can mitigate more memorization samples without lost of information from prompts. This ablation study implies the benefits of combing the two components of our method.

%% file: secs/append.tex
\newpage
\appendix

\section{Additional Experiments}
\label{append:sd2}

\subsection{Further Verification of Memorization's Reliance on the Embeddings of Trigger Tokens}
\label{exp:trigger_test}

\begin{figure}[tb]
  \centering
  \begin{subfigure}{0.48\linewidth}
  \centering
    \includegraphics[width=\textwidth]{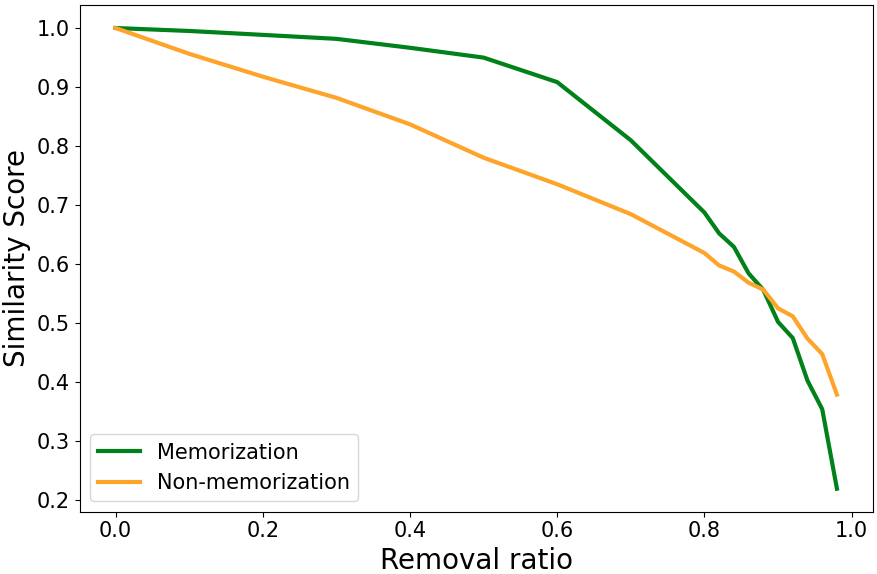}
    \caption{}
    \label{fig:trigger_test}
  \end{subfigure}
  \begin{subfigure}{0.48\linewidth}
  \centering
    \includegraphics[width=\textwidth]{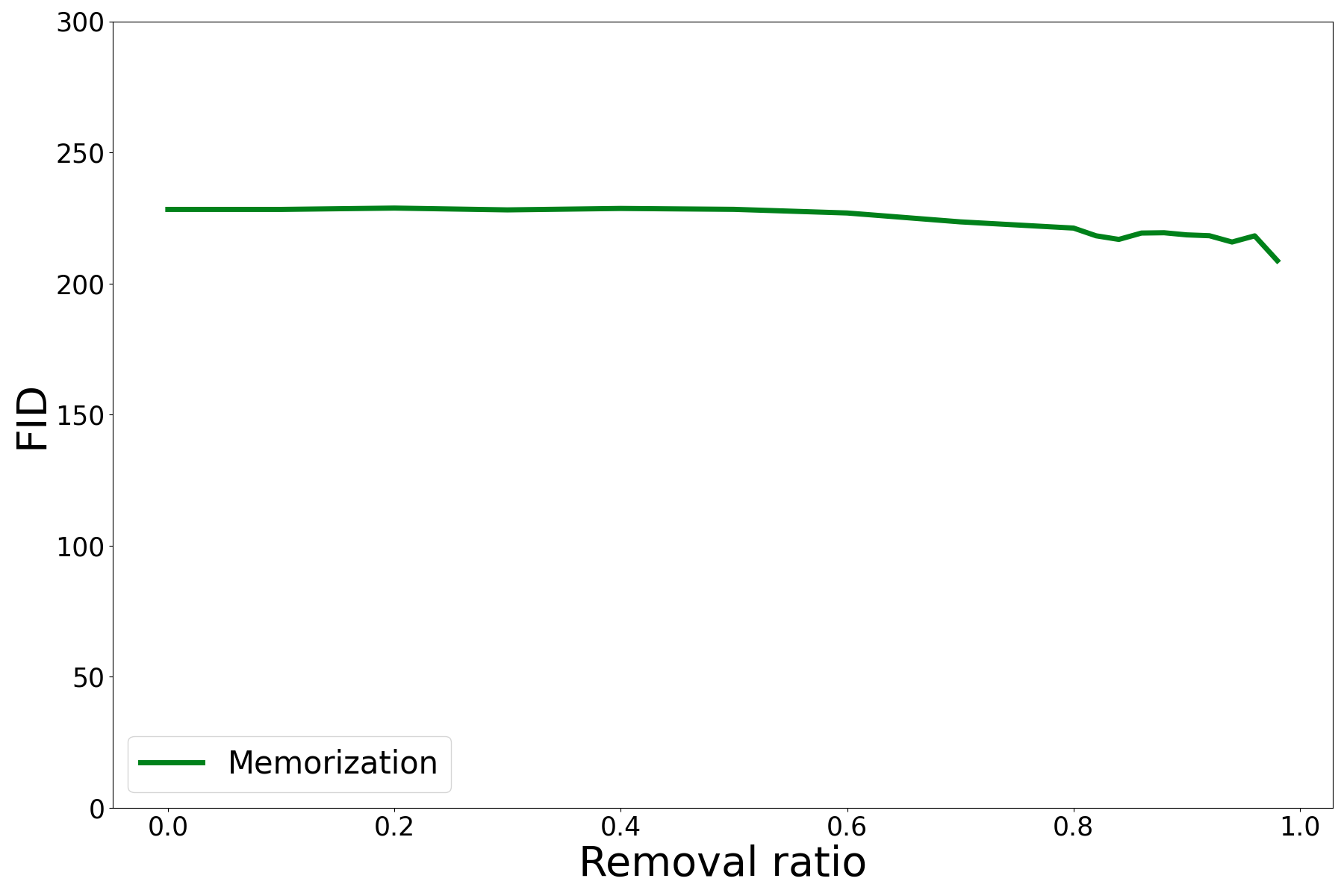}
    \caption{}
    \label{fig:trigger_test2}
  \end{subfigure}
  \caption{(a)Similarity score between generations with and without removing tokens; (b) FID between generations with and without removing tokens}
\end{figure}

In this subsection, we provide the experiment in \cref{fig:trigger_test} to further verify that the memorization relies on the existence of trigger token embeddings. While \cref{sec:property} compares the behavior of the attention score with/without memorization, in \cref{fig:trigger_test}, we add a mask to the attention scores to partially the trigger embeddings and examine the memorization. 

To be specific, we first generate images by a group of memorized prompts of SD v1.4 and a group of non-memorized prompts and save the random seed of this generation. Then we use the same prompts and seed to generate more rounds. For each round, we remove non-<begin> token embeddings one by one from small attention scores to large attention scores, {by masking out the logits of tokens}. Note that this does not influence the embeddings in $e_c$ and $V(e_c)$, and only changes whether the embedding of a token is used or not. 

Since memorization relies on trigger token embeddings and trigger tokens always hold large attention scores, it is expected that the generated images will not change much when only tokens with small attention are removed. This conjecture aligns with our experimental results: In \cref{fig:trigger_test}, we depict the similarity score between the generation with removed tokens and the first-round generation. When the removal ratio increases, the similarity score of non-memorization reduces much faster in the beginning, while memorized prompts have almost no reduction when even 60\% of the tokens with small attentions are removed. Instead, when the removal ratio is lager than 90\%, the similarity score of memorization reduces much faster than non-memorization, it means more trigger tokens are removed at this stage. Thus, we further confirm our observation that the memorization relies on the existence of trigger token embeddings.

\textit{Remark.} The reduction of similarity does not mean the reduction of generation quality. It only indicates the same seed can produce dissimilar images from the same prompt. In \cref{fig:trigger_test2}, we show that FID of memorization samples is even improved. This is similar to the reason we mentioned in \cref{exp:mitigation} that when memorization effect is mitigated, the generated images can become more diverse, which brings a reduced FID.


\subsection{Validation of Findings on SD v2.0}
\label{append:findings_sd2}

In this subsection, we implement the experiments to valid the generalization of our findings on SD v2.0. 

\subsubsection{Finding 1.} 
\begin{figure}[tb]
  \centering
  \includegraphics[width=0.5\textwidth]{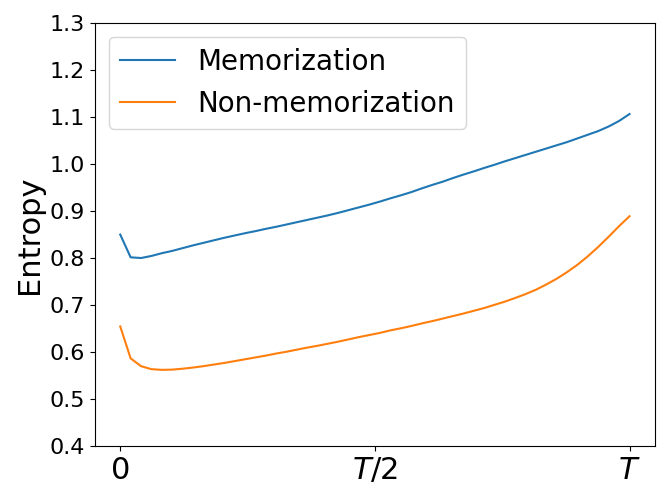}
  \caption{Entropy of SD v2.0}
  \label{fig:finding1_sd2}
\end{figure}
We plot the entropy of all the diffusion steps of SD v2.0 in \cref{fig:finding1_sd2}. It is shown that, for memorization samples, the entropy is much higher than non-memorization, which is consistent with our \textbf{Finding 1} that the trigger tokens will divert the attention from the beginning tokens, resulting in a more diverse attention distribution. Compared with SD v1.4, SD v2.0 has higher entropy even when $t$ is large. This means the concentration of attention on trigger tokens are more obvious than SD v1.4 when $t$ is large. 

\subsubsection{Finding 2.} 
\begin{figure}[tb]
  \centering
  \includegraphics[width=0.5\textwidth]{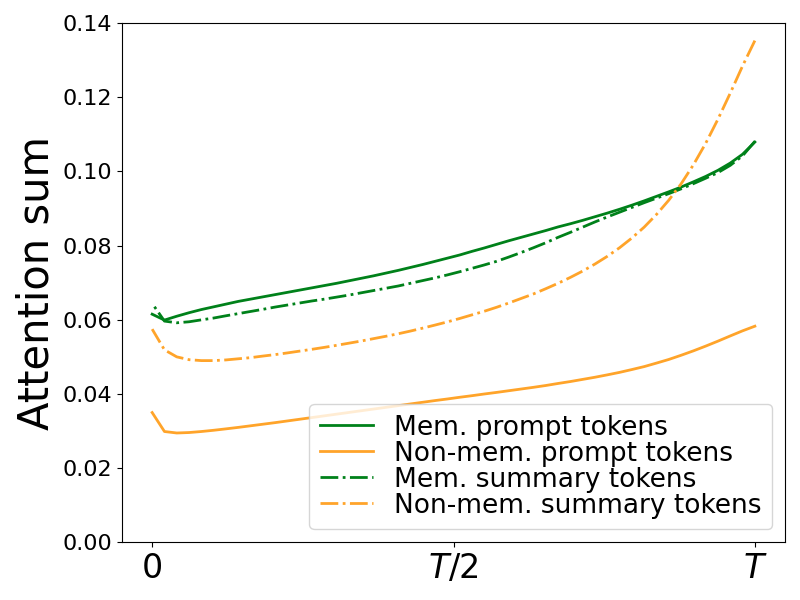}
  \caption{Attention sum of prompt tokens and summary tokens on SD v2.0}
  \label{fig:finding2_sd2}
\end{figure}
There are two observations in \textbf{Finding 2}. One observation is for different types of memorization and the other observation is focusing on a shared feature of all the types. Due to the fact that SD v2.0 is trained on a de-duplicated dataset, we can only observe TM and very few RM in it. Thus, we only validate the shared observation that the non-memorization will have a faster reduction of attention scores on summary tokens. We plot the attention sum of prompt tokens and summary tokens in \cref{fig:finding2_sd2}. It can be observed that the attention sum on the summary tokens of non-memorization samples reduces much faster than that of memorization samples. This phenomenon is used in the detection of memorization of SD v2.0.

\subsubsection{Finding 3.} 
\begin{figure}[tb]
  \centering
  \includegraphics[width=0.8\textwidth]{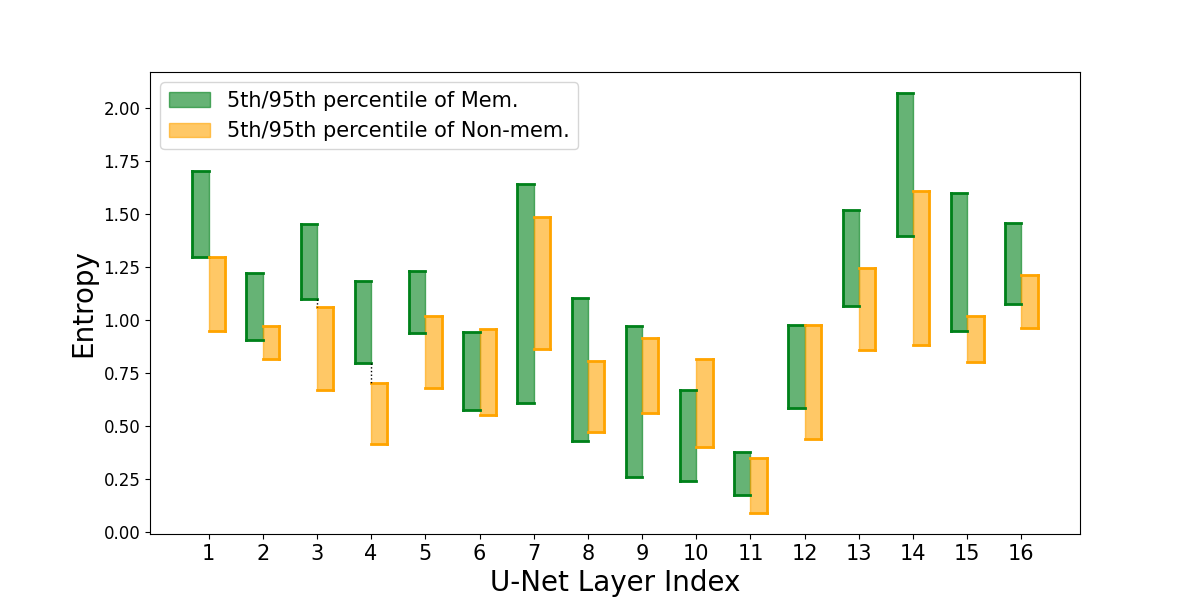}
  \caption{Entropy of each U-Net layer in SD v2.0}
  \label{fig:single_layer_95_sd2}
\end{figure}
To verify this finding on SD v2.0, we plot the entropy of each U-Net layer of SD v2.0 in \cref{fig:single_layer_95_sd2}. We can see that the overlapping of entropy in different layers are still different and diverse. The third and fourth layers have a clear separation.

In summary, all the findings in \cref{sec:property} can be generalized to SD v2.0. In \cref{exp:detect}, we also use these findings to detect the memorization on SD v2.0. These experiments shows the generalization of our findings to other text-to-image models with cross attention.

\subsection{Extension to T5-based and Transformer-based Diffusion Models}
\label{appd:transformer_t5}

In this subsection, we will provide results and analysis for T5-based models~\cite{chen2023pixart,deepif}, transformer-based models~\cite{chen2023pixart}, and SD fine-tuned for high quality \cite{epiCRealism,wallace2023diffusion}. We conduct experiments to verify that our detection and mitigation methods also work on these models, as in Tab.~\ref{tab:ext_detect} and Tab.~\ref{tab:ext_miti}.

\begin{table}[tb]
\centering
\caption{Extension of detection to other T2I models}
\label{tab:ext_detect}
\begin{tabular}{ccccc}
\toprule
& \multicolumn{2}{c}{Different Architecture} & \multicolumn{2}{c}{High-Quality Fine-tuning} \\
& ~~IF~\cite{deepif}~~ & ~Pixart-$\alpha$~\cite{chen2023pixart}~ & ~epiCRealism~\cite{epiCRealism}~ & ~Diffusion-DPO~\cite{wallace2023diffusion}~ \\
\midrule
AUROC & 0.929 & 0.903 & 0.905 & 0.958 \\
\bottomrule
\end{tabular}
\end{table}

\begin{table}[tb]
\centering
\caption{Extension of mitigation to other T2I models}
\label{tab:ext_miti}
\begin{tabular}{ccc}
\toprule
& ~w/ mitigation~ & ~w/o mitigation~ \\
\midrule
epiCRealism & 0.24 & 0.59 \\
Diffusion-DPO & 0.30 & 0.68 \\
\bottomrule
\end{tabular}
\end{table}

To summarize the \textit{experiments}, for detection in Tab.~\ref{tab:ext_detect} (following settings in \cref{tab:detection} in the main paper), our method can achieve AUROC higher than 0.9 for all new models. For mitigation in Tab.~\ref{tab:ext_miti} (following settings in \cref{fig:infer_miti} in the main paper), the inference-time mitigation reduces the similarity significantly to 0.24 and 0.30 on~\cite{epiCRealism} and \cite{wallace2023diffusion}.

To \textit{explain} the behavior of the models in detail, while the high-level intuition of concentrated attention on trigger tokens in memorization holds, there are some slight differences among different models. For T5-based, T5 encoder is bi-directional, which is different from the causal CLIP encoder in Stable Diffusion (SD). This means that \textit{end} and \textit{padding} tokens do not necessarily summarize the whole sentence. Besides, T5 encoder has no \textit{beginning} tokens, while SD concentrates attention on the CLIP \textit{beginning} token when focusing on denoising. Thus, {Pixart-$\alpha$}~\cite{chen2023pixart}, which is T5- and transformer-based, concentrates on the \textit{end} token (This is possibly because \textit{end} token contains less information and Pixart-$\alpha$ knows \textit{end} token). The trigger token will distract attention from \textit{end} token, which leads to disperse distribution and a high entropy. When finally using the entropy to detect memorization, we can still observe the discrepancy with/without memorization. The detection performance on Pixart-$\alpha$ shows that our main findings still holds on T5-based and transformer-based diffusion models.

For {ImageFloyd (IF)}~\cite{deepif}, it is special because the cross attention is built between the image and the concatenation of prompt tokens and the image itself. The attention is in a shape of $D_I \times (D_t + D_I)$, where $D_I$ is image dimension, and $D_t$ is prompt dimension. In the steps focusing on denoising, the image distributes most of attention to the image itself. (The image covers most of attention, while each dimension of image has a small attention.) Thus, for memorized prompts, attention will concentrate on trigger tokens with no competitor token. This leads to a low entropy in contrast to all other models.

For fine-tuned SD for high-quality, the aforementioned \cite{epiCRealism} and \cite{wallace2023diffusion} are fined-tuned from checkpoint of SD. We find that they also memorize most of prompts of SD.

\section{Entropy of Summary Tokens}
\label{append:summary_entropy}

\begin{figure}[tb]
  \centering
  \begin{subfigure}{0.32\linewidth}
  \centering
    \includegraphics[width=0.90\textwidth]{figs/padding_sd1_mv.png}
    \caption{MM}
    \label{fig:attn_sum_mv_appd}
  \end{subfigure}
  \begin{subfigure}{0.32\linewidth}
  \centering
    \includegraphics[width=0.90\textwidth]{figs/padding_sd1_rv.png}
    \caption{RM}
    \label{fig:attn_sum_rv_appd}
  \end{subfigure}
  \begin{subfigure}{0.32\linewidth}
  \centering
    \includegraphics[width=0.90\textwidth]{figs/padding_sd1_tv.png}
    \caption{TM}
    \label{fig:attn_sum_tv_appd}
  \end{subfigure}

  \begin{subfigure}{0.32\linewidth}
  \centering
    \includegraphics[width=0.90\textwidth]{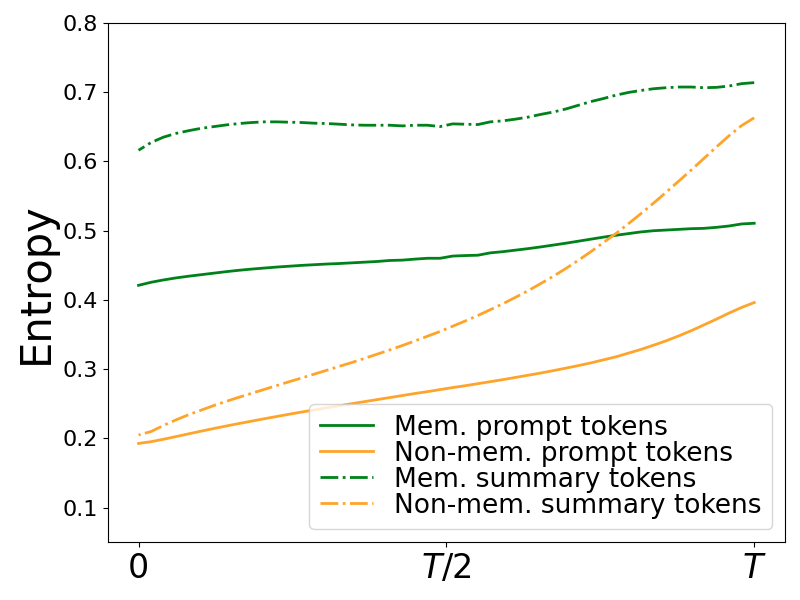}
    \caption{MM}
    \label{fig:ex_mm_appd}
  \end{subfigure}
  \begin{subfigure}{0.32\linewidth}
  \centering
    \includegraphics[width=0.90\textwidth]{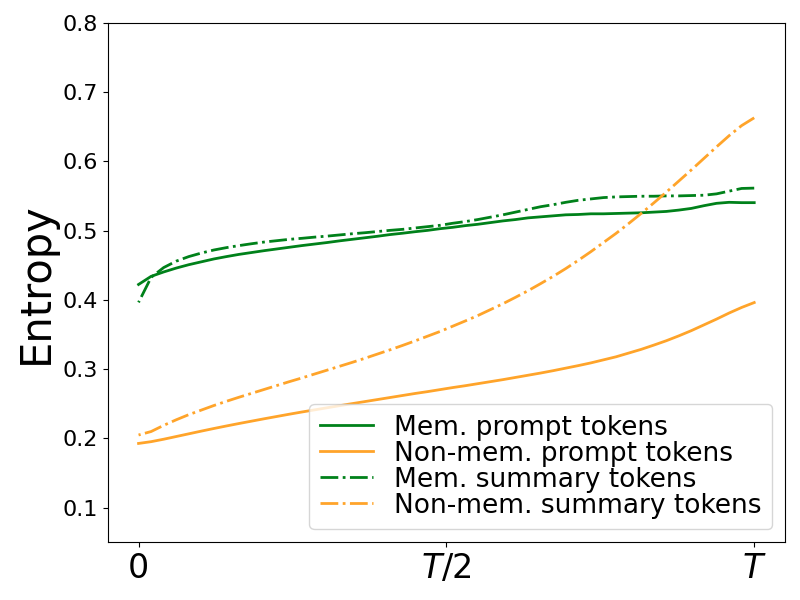}
    \caption{RM}
    \label{fig:ex_rm_appd}
  \end{subfigure}
  \begin{subfigure}{0.32\linewidth}
  \centering
    \includegraphics[width=0.90\textwidth]{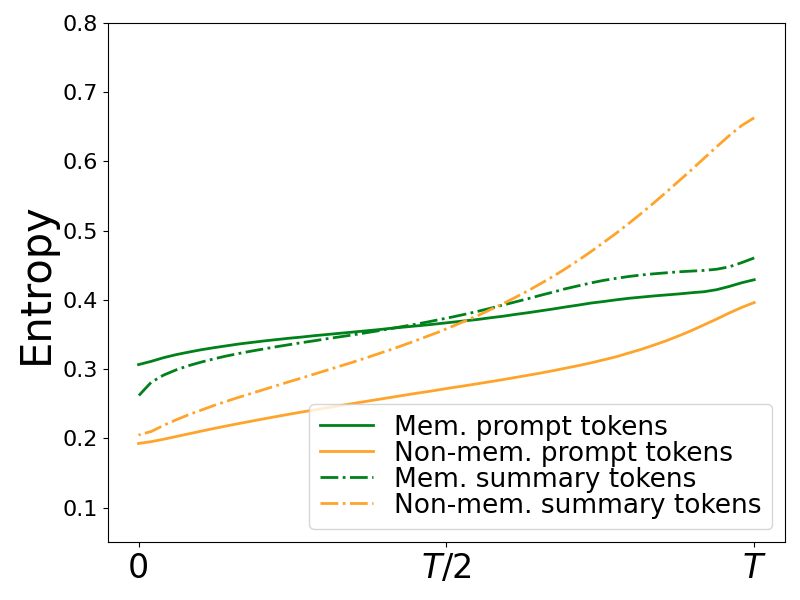}
    \caption{TM}
    \label{fig:ex_tm_appd}
  \end{subfigure}
  \caption{(a)(b)(c) Sum of attention score on prompt tokens and summary tokens in MM, RM and TM (the same results as \cref{fig:three_types}); (d)(e)(f) The entropy calculated on prompt tokens and summary tokens, \ie $E_t^{\text{prompt}}$ and $E_t^{\text{summary}}$, of MM, RM and TM.}
  \label{fig:three_types_appd}
\end{figure}
In \cref{sec:detect}, we use $E_t^{\text{summary}} = \sum_{i=N-S+1}^N-\bar{a}_i \log \left(\bar{a}_i\right)$ to replace sum to make its numerical values comparable with $E_t$ in \cref{eq:measure1}. In this section, we show that $E_t^{\text{summary}}$ has the consistent conclusion with \textbf{Finding 2}, thus it can be used for replacing sum in \textbf{Finding 2}.

In \cref{fig:three_types_appd}, we show both the attention sum and $E_t^{\text{prompt}}$ on three types of memorization. For the first observation of \textbf{Finding 2}, we can see that for MM, the attention sum on summary tokens is larger than prompt tokens, while the attention sums on summary tokens of RM and TM are lower than prompt tokens. This difference between MM and the other two types is explained in \cref{fig:three_types_appd}, that MM tend to have more unique prompts and paying higher attention on the summary tokens is easier to memorize the prompts.  As for $E_t^{\text{prompt}}$ and $E_t^{\text{summary}}$, we can see that MM have much higher $E_t^{\text{summary}}$ than $E_t^{\text{prompt}}$, while in RM and TM, $E_t^{\text{summary}}$ is almost similar to $E_t^{\text{prompt}}$. Although it is different from the attention sum that the summary tokens in RM and TM are smaller than prompt tokens, it is enough to show the different patterns of between MM and the other two types of memorization.

For the second, observation of \textbf{Finding 2}, we can see that, in all three types of memorization, both attention sum and entropy of summary tokens of non-memorization reduce faster than those of memorization. This property is used in the first detection metric in \cref{eq:measure1}.

In summary, $E_t^{\text{summary}}$ is consistent with the attention sum in the two observations in \textbf{Finding 2}, thus this is reasonable to replace the sum with $E_t^{\text{summary}}$ in \cref{eq:measure1} to make the numerical values comparable with the first term. 

\section{Details of Experimental Settings}
\label{append:detail_experiment}

\subsection{Baselines}
\label{append:baselines}

\subsubsection{Detection.}

For the baseline method from Carlini \etal~\cite{carlini2023extracting}, we generate $n$ samples with different random seeds for each prompt, and calculate the $l_2$ distance between them in pairs. We use the smallest distance among the pairs as the measurement. Since the memorization samples will produce the same results no matter what the random seed is, the $l_2$ distance between generated images of memorization should be small, while the non-memorization samples should have larger distance. The weakness of this method is that it requires repeated generation for one prompt, which is slow.

For the baseline method from Wen \etal~\cite{wen2023detecting}, we compare the difference of the model output between the suspect prompt and empty prompt. To stabilize the results, we generate $n$ samples on $s$ diffusion steps, and use the averaged results for detection. This method will also slow down the generation speed because it needs additional inference operation on empty prompt, and may use repeated generation for stabilized results.

\subsubsection{Mitigation.}

Wen \etal~\cite{wen2023detecting} provide two variants of mitigation methods for inference and training. For inference-time mitigation, they update the initial token embeddings of the prompt to make the model output closer to the empty prompt. For training-time mitigation, they set up a threshold to remove the samples with high difference from empty prompt. This also requires additional inference operation on the empty prompt. In \cref{append:train_threshold}, we compare how to select the threshold between their method and ours.

\subsection{Metrics}
\label{append:metrics}

\subsubsection{Detection.} We calculate the time cost on a single GPU of A5000 with batchsize of 32 for SD v1.4 and 16 for SD v2.0.

\subsubsection{Mitigation.} Similarity Score is calculated between the generated images and the training images paired with memorized prompts. For inference-time mitigation, the FID is calculated between generated images (including both memorized prompts from~\cite{webster2023reproducible} and 500 non-memorized prompts) and a subset of LAION~\cite{schuhmann2022laion} with 10,000 images. For training-time mitigation, CLIP score is calculated on 100 memorized prompts in training data, 200 non-memorized prompts in training data and 200 generated prompts by ChatGPT-4.

\subsection{Training-time mitigation threshold}
\label{append:train_threshold}

For both baseline method and ours, we need a threshold to remove the samples from the training mini-batch. Based on the extracted memorized samples in \cite{webster2023reproducible}, we use the threshold of the best F1 score for the baseline method. To control the strength as shown in \cref{fig:train_miti}, we adjust the threshold by adding a number in the range of -2 to 1. For ours, since we have noticed that different diffusion steps tend to have different entropy, we use each step's 5th-percentile of the entropy of memorized samples in \cite{webster2023reproducible} as the threshold for that diffusion step. This can remove the samples more accurately based on each step. For larger $t$, the threshold is larger since the entropy is larger in the early stage of the diffusion process; conversely, for smaller $t$, the threshold is smaller. To control the strength, we adjust the threshold by multiplying it with a factor. In our experiment, we set the factor in the range of 1.1 to 1.35.